\documentclass[conference]{IEEEtran}
\IEEEoverridecommandlockouts
\usepackage{cite}
\usepackage{amsmath,amssymb,amsfonts}
\usepackage{graphicx}
\usepackage{textcomp}
\usepackage{xcolor}
\usepackage{algorithm}
\usepackage{multirow}
\usepackage{times}
\usepackage{helvet}
\usepackage{courier}
\usepackage{algpseudocode}
\usepackage{algorithm}
\usepackage{amsfonts}
\usepackage{diagbox}
\usepackage{lineno}
\usepackage{subfigure}
\usepackage{url}
\usepackage{lipsum}

\def\BibTeX{{\rm B\kern-.05em{\sc i\kern-.025em b}\kern-.08em
    T\kern-.1667em\lower.7ex\hbox{E}\kern-.125emX}}
\begin{document}

\title{IFGAN: Missing Value Imputation using Feature-specific Generative Adversarial Networks}

\author{\IEEEauthorblockN{Wei Qiu*}
\IEEEauthorblockA{\textit{Yuanpei College} \\
\textit{Peking University}\\
Beijing, China \\
qiuweipku@gmail.com}
\and
\IEEEauthorblockN{Yangsibo Huang*}
\IEEEauthorblockA{\textit{College of Computer Science and Technology} \\
\textit{Zhejiang University}\\
Hangzhou, China \\
hazelsuko@gmail.com}
\and
\IEEEauthorblockN{Quanzheng Li}
\IEEEauthorblockA{\textit{Department of Radiology} \\
\textit{Massachusetts General Hospital}\\
Boston, MA \\
li.quanzheng@mgh.harvard.edu}
}

\maketitle

\begin{abstract}
Missing value imputation is a challenging and well-researched topic in data mining. In this paper, we propose IFGAN, a missing value imputation algorithm based on Feature-specific Generative Adversarial Networks (GAN). Our idea is intuitive yet effective: a feature-specific generator is trained to impute missing values, while a discriminator is expected to distinguish the imputed values from observed ones. The proposed architecture is capable of handling different data types, data distributions, missing mechanisms, and missing rates. It also improves post-imputation analysis by preserving inter-feature correlations. We empirically show on several real-life datasets that IFGAN outperforms current state-of-the-art algorithm under various missing conditions.
\end{abstract}


\section{Introduction}
\footnotetext{* Joint first authors}
Missing values imputation is crucial in data mining, especially for large-scale healthcare\cite{mimic, verma2014imputation} and social\cite{davern2004missing, liu2015missing} studies, where missingness is a common issue. Data can be missing for various reasons: different subjects may have different requested measurements; errors in the data collecting device make certain data entries unavailable; or subjects may refuse to answer some items from a survey. On top of the missing mechanism, datasets with missing can also vary in the data type (i.e., continuous, categorical, or mixed-type), sample size and feature size.  A good imputation algorithm is expected to generalize to a diversity of missing conditions mentioned above, by estimating missing values based on the observed data.

However, as we will demonstrate in detail in the next section, classical imputation methods have different limitations: for instance,  when it comes to missing data type, $k$ nearest neighbours \cite{troyanskaya2001missing} is only capable of imputing continuous data, and speaking of missing rate, matrix completion models like Singular Value Decomposition (SVD) \cite{mazumder2010spectral}  may not perform well on heavily missing dataframes, such as electronic health records. Besides, some models have certain distributional assumptions (e.g. assuming normal distributions) and hence cannot handle arbitrary missing patterns.

In contrast to classical models, deep learning, as one of the most currently remarkable machine learning techniques, has established state-of-the-art performance in many applications. When it comes to the task of imputation, deep architectures have also shown greater promise than classical ones in recent studies\cite{angermueller2017deepcpg, yoon2018gain, gondara2018mida}, due to their capability to automatically learn latent representations and inter-variable associations. Nonetheless, these work all formulated data imputation as a mapping from a missing matrix to a complete one, which ignores the difference in data type and statistical importance between features and makes the model unable to learn feature-specific representations.

To overcome the limitations of classical models and to incorporate feature-specific representations into deep learning frameworks, we propose IFGAN, a feature-specific generative adversarial network for missing data imputation. 

The contribution of this paper is fourfold:
\begin{itemize}
    \item We introduce a feature-specific deep adversarial imputation pipeline, which is capable of accurately imputing any type of input data, i.e., continuous, categorical or mixed. The training process helps preserve inter-feature correlations, which also improves post-imputation analysis such as classification. 
    \item  We empirically demonstrate that taking benefits of adversarial training, our approach fills in unobserved entries with more `realistic' values, namely more compliant with original data distribution, than other baselines.
    \item We further evaluate the proposed method under three different missing mechanisms, where our model show state-of-the-art performance in all scenarios, even for the missing not at random(MNAR) case, which is a performance bottleneck for most imputation methods. 
    \item We demonstrate that our model can train with a diverse range of data availability, including various missing rates, sample size, and feature size. This property enables our method's application in diversified real-life situations.
\end{itemize}


\section{Related Work}
\label{sec_related}
Missing value imputation is a challenging and well-researched topic in statistics because the missingness of dataset may cause bias and affect the quality of subsequent data mining processes. Most mainstream approaches firstly capture statistical properties of a given dataset with missingness, and then utilize a pre-defined model for its imputation.


Although there are plenty of model-based imputation methods, many of them have limitations on the input dataset. Some methods can only impute one type of missingness, for example, $k$ nearest neighbours (KNN)\cite{troyanskaya2001missing} and matrix completion methods\cite{troyanskaya2001missing, mazumder2010spectral}, can only be used to impute continuous data. Other approaches may pose assumptions about the underlying statistical properties of data. For instance, the imputation method based on Expectation Maximization\cite{garcia2010pattern} assumes input vectors to be generated following a probability density function; multivariate imputation by chained equations (MICE)\cite{azur2011multiple} depends on a specific parametric model which makes assumptions about the distribution of the data, and can lead to questionable situation when the dataset can't satisfy the assumption.

To overcome the limitation introduced by the assumptions on input data, recently proposed methods turn to deep learning architectures. However, deep models usually need an explicitly defined training set to optimize on, which is almost impossible to acquire for our data imputation problem. An alternative is to train the model only with observed components, as the Denoising AutoEncoder model attempted\cite{gondara2018mida}. However,  in that case, the number of observed entries may be insufficient for the model to learn a good representation. Moreover, these deep models usually treat the whole data frame as input and train a dataset-specific global model, which unintendedly ignore feature-specific properties and information.

Another line of work worth noticing is the image completion task, where most current state-of-the-art approaches adopt a Generative Adversarial Networks based architecture since it was proposed\cite{pathak2016context,iizuka2017globally, liu2018image}. Although there exist differences between the input images in their work and data frames in our case, they share more similarities. Without losing generality, the image completion task can be formulated as a specific data frame imputation problem, where each entry of the data frame is a pixel value in the image. For GAN-based image completion, the objective of the generator is to output pixel values to complete the corrupted image, and the discriminator aims to tell which pixels are synthetic. Inspired by this, we adopt GAN for data imputation in a similar way in our work, where the generator works to guess missing values and the discriminator is trained to tell the dissent between imputed values and observed ones.

The closest work to ours is GAIN\cite{yoon2018gain}, which also uses a GAN-based architecture for imputation. However, their work is almost identical to an image completion pipeline except that the input is a data frame with missingness instead of a corrupted image. They failed to train different models for different features, and the network mixed up categorical and continuous variables in training. Our approach, to the contrary, is customized to the data imputation task by adopting a feature-specific optimization scheme. We will discuss in detail our algorithm and demonstrate its effectiveness in the following sections.


\begin{table*}
\centering 
\setlength{\tabcolsep}{10mm}
\caption{Summary of Notations.} \label{not} 
\begin{tabular}{c|c} 
\hline 
\textbf{Symbol} &\textbf{Definition} \\
\hline 
\hline
$\textbf{X}^{com}$ & A data matrix without missing values\\
\hline
\multirow{2}{*}{$\textbf{M}$} & A mask matrix taking values in \{0, 1\} \\&that indicates which components of $\textbf{X}$ are observed\\
\hline
\multirow{2}{*}{$\hat{\textbf{X}}$} &A data matrix with missing values, \\
& which can be manually constructed by
$\textbf{X}^{com}\cdot\textbf{M}$\\ 
\hline
\multirow{2}{*}{$\textbf{X}$} & Filling the missing values in $\hat{\textbf{X}}$ \\ & with the imputed values\\
\hline
$\textbf{X}_{i}$ & The $i$-th column of data matrix $\textbf{X}$\\
\hline
\multirow{2}{*}{$\textbf{X}_{\sim i}$} &  The columns of data matrix $\textbf{X}$ \\
&other than the $i$-th column\\
\hline
$\textbf{X}(j)$& The $j$-the row of data matrix $\textbf{x}$\\
\hline
$\textbf{Z}$ & A uniform random matrix with the same size as \textbf{X}\\
\hline
$\textbf{1}$ & An all-ones matrix \\
\hline
$G$ & A group of imputation generators\\
\hline
$D$ & A group of imputation discriminators\\
\hline
$G_i$ & A feature specific-generator for the $i$-th feature\\
\hline
$D_i$ & A feature specific-discriminator for the $i$-th feature\\
\hline
$N$ & The number of samples in the dataset.\\
\hline
$d$ & The number of features in the dataset.\\

\hline 
\end{tabular} 
\end{table*}

\section{Problem Formulation}
\label{sec_pro}
Throughout this paper, we use the capital alphabet in boldface, $e.g.$ $\textbf{X}$, to denote a matrix or a dataset. We use the capital alphabet in boldface with subscript to denote a single or a group of column-based vector(s), $e.g.$ $\textbf{X}_{i}$ refers to the $i$-th column of $\textbf{X}$ and $\textbf{X}_{\sim i}$ refers to the columns of $\textbf{X}$ other than the $i$-th column. For clarification, we summarize important notations used by this paper in Table \ref{not}.

In order to impute an arbitrary column $\textbf{X}_i$ in data matrix $\textbf{X}$, we can firstly separate $\textbf{X}$ in four parts:
\begin{enumerate}
    \item The observed values of $\textbf{X}_i$, denoted by $\textbf{X}_i^{obs}$;
    \item The missing values of $\textbf{X}_i$, denoted by $\textbf{X}_i^{mis}$;
    \item The columns other than $\textbf{X}_i$ within the samples which has observed value in $\textbf{X}_i$, denoted by $\textbf{X}_{\sim i}^{obs}$;
     \item The columns other than $\textbf{X}_i$ within the samples which has missing value in $\textbf{X}_i$, denoted by $\textbf{X}_{\sim i}^{mis}$.
\end{enumerate}

\begin{figure}
    \centering
    \includegraphics[width=3.5in]{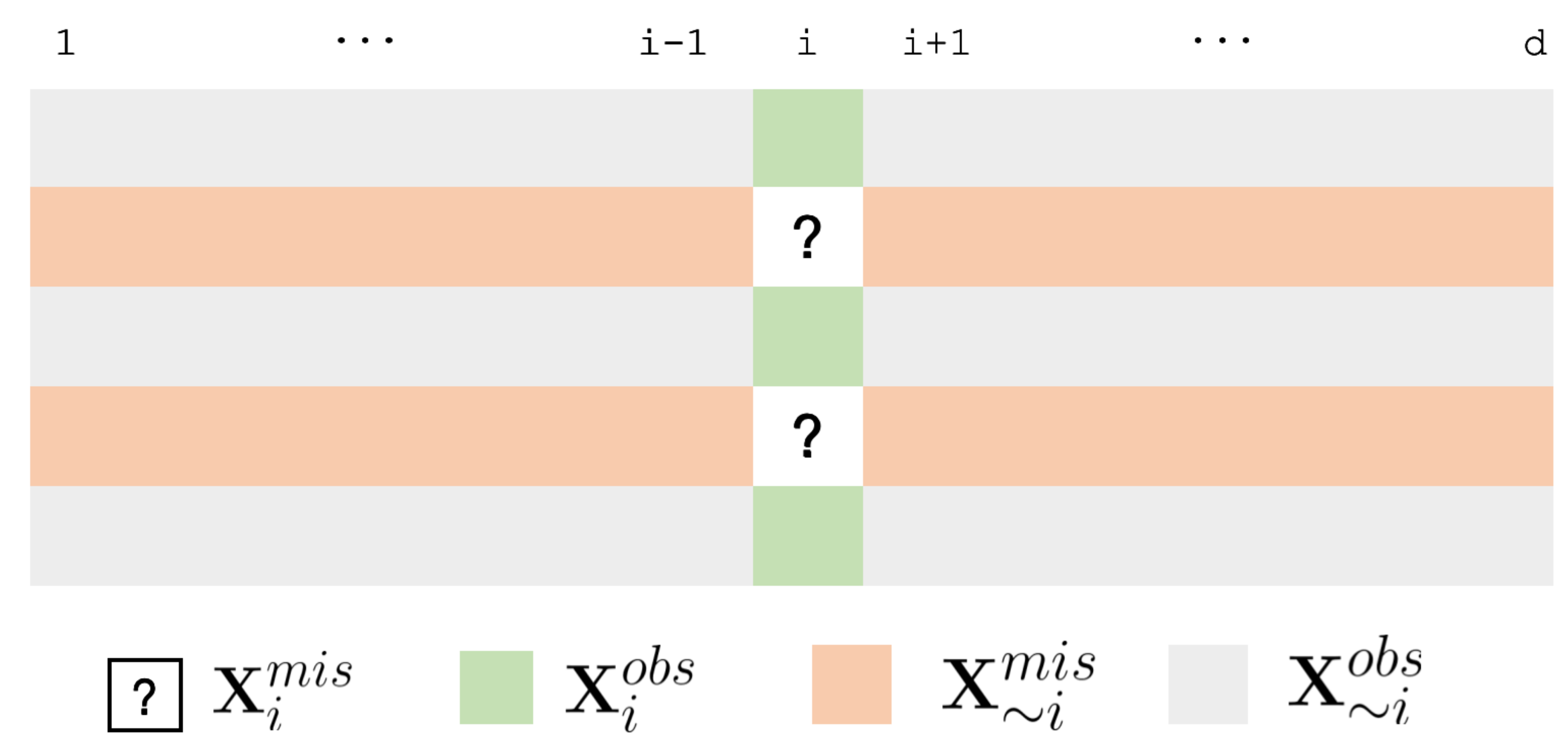}
    \caption{Our separation of a data matrix to be imputed.}
    \label{fig:4part}
\end{figure}

For better illustration, we visualize these four parts in Figure \ref{fig:4part}. The goal of imputation is to minimize the distribution difference between the imputed values $\textbf{X}^{mis}_i$ and observed ones $\textbf{X}^{obs}_{i}$. To achieve this, we use a multiple imputation framework to generate multiple draws, which can model the distribution of the data rather than just the expectation. 

When it comes to the prior distribution of missing data, we take into consideration the following three missing mechanisms. To account for these three missing mechanisms, we use an example of questionnaires about income within a certain population. 
\begin{enumerate}
    \item \textbf{MCAR.} The data is missing completely at random (MCAR) if the missingness occurs entirely at random. For example, the survey participants flip a coin to decide whether to answer questions or not. 
    \item \textbf{MAR.} The data is missing at random (MAR) if the missingness depends only on the observed variables. For example, the questionnaires include the profession subject, and participants whose profession is manager are more likely not to share their income. In this case, the missingness in `income' is MAR.
    \item  \textbf{MNAR.} The data is missing not at random (MNAR) if the missingness depends on the missing values themselves. For example, suppose the participants don't want to share their income when it is below 1000 dollars per month. In this case, the missingness in `income' is MNAR.
\end{enumerate}

Our feature-specific imputation algorithm is designed to be robust to all three missing mechanisms mentioned above. To prove this, we provide results for our algorithm and compare them with other benchmarks under these mechanisms in the Experiment section.

\section{Methodology}
\label{sec_method}

\begin{figure*}[!h]
\label{fig_workflow}
\centering
\caption{The workflow of IFGAN. The discriminator $D$ consists of $d$ feature-specific discriminators $D_i$, where $d$ is the number of features. Each $D_i$ takes $\textbf{X}$ as input, and outputs a vector $D(\textbf{X})$ $\in (0,1)^N$, where the $i$-th component of $D_i(\textbf{X})$ denotes the probability that $\textbf{X}_i$ is missing. The ground truth value of $D_i(\textbf{X})$ is $\textbf{1}-\textbf{M}_i$. The loss function indicates the difference between the output and $\textbf{1}-\textbf{M}_i$. The generator $G$ is a combination of $d$ feature-specific generators $G_i$. Each $G_i$ works as the imputer for the $i$-th feature. To be specific, each $G_i$ takes $\textbf{X}_i^{obs}$, $\textbf{X}_{\sim i}^{obs}$, $\textbf{X}_{\sim i}^{mis}$ as input, and outputs $\textbf{X}_i^{mis}$, a completion for the missing values in the $i$-th column of $\hat{\textbf{X}}$. The loss function is the difference between the output and $\textbf{X}_i^{obs}$ minus the discriminators $D_i$'s loss.}
\includegraphics[width=7in]{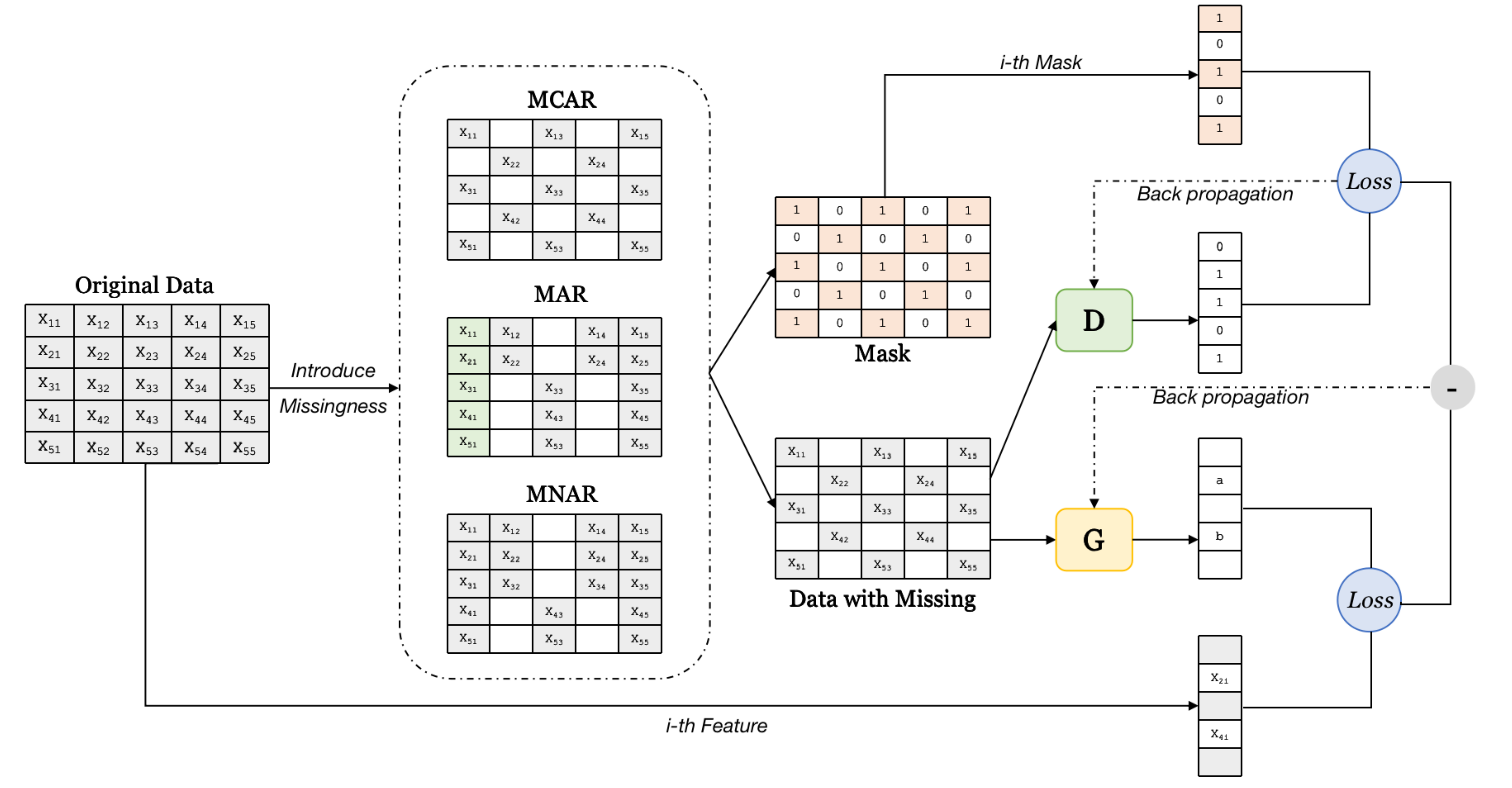}
\end{figure*}

In this section, we introduce the detailed design of our
proposed Imputation algorithm, Feature-specific Generative Adversarial Networks (IFGAN). The overall architecture of the proposed method shows in Figure 2. We first briefly review the key components of conventional GANs, including its objective function and optimization approach. Then we present the key idea of generators in IFGAN by illustrating how they make feature-specific predictions for missing values. Afterward, we discuss the tailored column-specific discriminators we used for IFGAN. Finally, we introduce our optimization strategy, which enables a robust adversarial training process.

\subsection{Generative Adversarial Network}
\label{GAN}
GAN\cite{goodfellow2014generative} is a group of generative models estimated via an adversarial process, in which a generator $G$ and a discriminator $D$ are simultaneously trained. The goal of GAN is to play the minimax game with value function $V (G, D)$:
\begin{equation}
\begin{aligned}
\min _{G} \max _{D} V(D, G)=\mathbb{E}_{\boldsymbol{x} \sim p_{\text { data }}(\boldsymbol{x})}[\log D(\boldsymbol{x})]\\
+\mathbb{E}_{\boldsymbol{z} \sim p_{\boldsymbol{z}}}(\boldsymbol{z})[\log (1-D(G(\boldsymbol{z})))]
\end{aligned}
\end{equation}
where $z$ is the noise which generator takes as input, $p_{\boldsymbol{z}}(\boldsymbol{z})$ represents the distribution of generated data, and $p_{data(x)}$ is the real-data distribution. Through this process, the generator is capable of generating more realistic samples, namely samples which more likely came from $p_{data(x)}$ rather than $p_{\boldsymbol{z}}(\boldsymbol{z})$.

We now revisit the GAN architecture in the setting of missing value imputation, where $Z$ represents data with missing entries, $G$ is the imputer which fills in missingness based on certain prediction algorithm, and $D$ aims to tell whether an entry is synthetic or from the original data frame. By obtaining the equilibrium of $D$ and $G$, the generator learns to fill in unobserved entries with values more compliant with the original data distribution. 

\subsection{Feature-specific Generator}
\label{Gen}
In order to deal with any type of input data, and to preserve correlations between features, we define the generators in a feature-specific manner. To be specific, our generator $G$ is a combination of $d$ feature-specific generators $G_i$, where $d$ is the number of features, and each $G_i$ works as the imputer for the $i$-th feature. Each $G_i$ takes $\textbf{X}_i^{obs}$, $\textbf{X}_{\sim i}^{obs}$, $\textbf{X}_{\sim i}^{mis}$ as input, and outputs $\textbf{X}_i^{mis}$, a completion for the missing values in the $i$-th column of $\hat{\textbf{X}}$.

Here we explain how $G_i$ learns the mapping from $\textbf{X}_i^{obs}$, $\textbf{X}_{\sim i}^{obs}$, $\textbf{X}_{\sim i}^{mis}$ to $\textbf{X}_i^{mis}$. Firstly, we train $G_i$ on the observed samples to fit $\textbf{X}_i^{obs} \sim \textbf{X}_{\sim i}^{obs}$ with the loss function:

\begin{equation}
\begin{aligned}
    \mathcal{L}_G(\textbf{X}_i^{obs},G_i(\textbf{X}^{obs}_{\sim i}))= \sum_{j=1}^{n} L_G(x_j,{x'_j}) +\lambda  \left\|w_{G_i}\right\|_2^2 
\end{aligned}
\end{equation}

where $G_i(\textbf{X}^{obs}_{\sim i})$ is the predicted vector of $\textbf{X}_i^{obs}$, $x_j$ and $x'_j$ are elements of $\textbf{X}_i^{obs}$, $G_i(\textbf{X}^{obs}_{\sim i})$ respectively, and $n$ being the number of the observed samples in the $i$-th column. $w_{G_i}$ stands for weight that $G_i$ learns, and $\lambda$ is the coefficient for L2 regularization. In order to make our model robust to different variable types, we define the element-wise loss $L_G$ as follow:

\begin{equation}
    L_G(x,{x'}) = \left\{
             \begin{array}{lr}
             -xlog(x') - (1-x)log(1-x'), &  if \ x \ is \ binary\\
             (x'-x)^2,  &  if \ x \ is \ continuous
             \end{array}
\right.
\end{equation}
Then the trained model takes $\textbf{X}_{\sim i}^{mis}$ as input to infer

\begin{equation}
\begin{aligned}
    \textbf{X}_i^{mis} = G_i(\textbf{X}_{\sim i}^{mis})
\end{aligned}
\end{equation}

By filling back $ \textbf{X}_i^{mis}$ to corresponding entries, we complete the task of imputation. Note that while \cite{yoon2018gain}'s generator outputs a value for every component, ours only need to output the values of missing entry, which greatly cuts down on the computation cost. 

\subsection{Feature-specific Discriminator} 
\label{Dis}
Merely using the generator for data imputation is enough to obtain imputation results comparable to the state-of-the-art, which we will illustrate in the Experiment section. However, we also design a simultaneously optimized discriminator to take benefit of adversarial training, which enables our model to impute unobserved entries with values more compliant with original data distribution.

Similar to our generator $G$, the discriminator $D$ also consists of $d$ feature-specific discriminators $D_i$. Each $D_i$ takes $\textbf{X}$ as input, and outputs a vector $D(\textbf{X})$ $\in (0,1)^N$, where the $i$-th component of $D_i(\textbf{X})$ denotes the probability that $\textbf{X}_i$ is missing. The ground truth value of $D_i(\textbf{X})$ is $\textbf{1}-\textbf{M}_i$.

Let $m'_j$ and $m_j$ be corresponding elements in vectors $D_i(\textbf{X})$ and $\textbf{1}-\textbf{M}_i$, then the loss of discriminator $\mathcal{L}_D$ is

\begin{equation}
\begin{aligned}
    \mathcal{L}_D(D_i(\textbf{X}),\textbf{1}-\textbf{M}_i) = \sum_{j=1}^N L_D(m_j,{m'_j}) +\lambda  \left\|w_{D_i}\right\|_2^2
\end{aligned}
\end{equation}

where $w_{D_i}$ is the weight $D_i$ learns, and $\lambda$ is the regularization coefficient. The element-wise loss $L_D$ is given by:

\begin{equation}
\begin{aligned}
    L_D(m_j,{m'_j}) = (m'_j-m_j)^2
\end{aligned}
\end{equation}

\subsection{Optimization algorithm}
\label{alg}
We solve the minimax problem in Equation 1 in an iterative manner. Both the generator $G$ and discriminator $D$ are modeled as multilayer perceptrons\cite{gardner1998artificial}.

The pseudo-code for the IFGAN algorithm is presented in Algorithm 1. The first step of our algorithm is making initial guesses for the missing values in $\hat{\textbf{X}}$ using random numbers. After this step, we can get the matrix $\textbf{X}$. Then, we sort the columns $\textbf{X}_{i}, i = 1,...,d$ according to the number of missing values in order from the lowest to highest\cite{kamishima2004filling}. For each column $\textbf{X}_{i}$, the missing values $\textbf{X}_{i}^{mis}$ are imputed by fitting our IFGAN architecture.

To be more specific, we alternately train the feature-specific generator $G_i$ to fit the mapping from $\textbf{X}_{\sim i}^{obs}$ to $\textbf{X}_i^{obs}$, and the discriminator $D_i$ to fit the mapping from $\textbf{X}$ to ($\textbf{1}-\textbf{M}_i$). Then, we predict the missing values $\hat{\textbf{X}}_{i}^{mis}$ by applying the trained generator to $\textbf{X}_{\sim i}^{obs}$ and fill them in the matrix. After one iteration of all columns, we can obtain $\textbf{X}^{old}$, which will be used in the next iteration. The imputation procedure is repeated until we meet the stopping criterion. 

Here we explain in detail how each $G_i$ and $D_i$ is trained and updated. Given a column $i$, we first optimize the discriminator $D_i$ with a mini-batch of size $b$. For each sample $\textbf{X}_{(j)}$ in the mini-batch, we predict the missing mask $D_i(\textbf{X}_{(j)})$ and calculate the difference between $D_i(\textbf{X}_{(j)})$ and the ground truth $1-\textbf{M}_{ij}$. We then update $D$ using the stochastic gradient decent algorithm\cite{bottou2010large} by:

\begin{equation}
\begin{aligned}
    \bigtriangledown_D  \sum_{j=1}^{b} \mathcal{L}_D(D_i(\textbf{X}_{(j)}),1-\textbf{M}_{ij})
\end{aligned}
\end{equation}

Once we have updated $D_i$, we optimize generator $G_i$ with the same mini-batch size $b$. While training $G_i$, we desire imputed values to 1) be as close as possible to corresponding true values in $\textbf{X}^{com}$, and 2) fool the discriminator, namely, make it hard to distinguish between a synthetic value and an observed one. To achieve both goals, we can design the update of the generator to be 
\begin{equation}
\begin{aligned}
    \bigtriangledown_G  (\sum_{j=1}^{b} \mathcal{L}_G(G_i(\textbf{X}^{obs}_{\sim ij}), \textbf{X}_{ij}^{obs}) - \alpha \mathcal{L}_D(D_i(\textbf{X}_{(j)},1-\textbf{M}_{ij})))
\end{aligned}
\end{equation}

where the first term guarantees the fidelity between the imputed and real values, and is minimized when the reconstructed data frame is close to the actually observed data. The second term is the adversarial term for the minimax game between $G_i$ and $D_i$.

\begin{algorithm}
\caption{Pseudo-code of IFGAN}
\begin{algorithmic}[1]
\label{alg 1}
  \Require
      An $N\times d$ data matrix with missing values $\hat{\textbf{X}}$, stopping criterion $\gamma$, max iteration $k$\\
      Make initial guess for missing values to get $\textbf{X}$;\\
      Sort indices of columns in $\textbf{X}$ by increasing amount of missing values, and store in $\textbf{C}$;
    \While{$\gamma$ is not met}
        \State $\textbf{X}^{old}$ $\leftarrow$ previously imputed matrix;
        \For{$c$ in $\textbf{C}$}
            \For{$j = 1,...,k$}
            \State Fit a generator $G_i$: $\textbf{X}^{obs}_{i}$ $\sim$ $\textbf{X}^{obs}_{\sim i}$
            \State Update $G_i$ by: 
            \State  \quad $ \bigtriangledown_G  (\sum_{j=1}^{b} \mathcal{L}_G(G_i(\textbf{X}^{obs}_{\sim ij}), \textbf{X}_{ij}^{obs})$
            \State \quad $ - \alpha \mathcal{L}_D(D_i(\textbf{X}_{(j)}),1-\textbf{M}_{ij}))$
            \State Predict and update $\textbf{X}^{mis}_{i}$: $\textbf{X}^{mis}_{i} = G_i(\textbf{X}^{mis}_{\sim i})$ ;
            \State Fit a discriminator $D_i$: $\textbf{1}-\textbf{M}_{i}$ $\sim$ $\textbf{X}^{old}$;
            \State Update $D_i$ by: 
            \State \quad $\bigtriangledown_D  \sum_{j=1}^{b} \mathcal{L}_D(D_i(\textbf{X}_{(j)}),1-\textbf{M}_{ij})$
            \EndFor
        \State $\textbf{X}^{new}$ $\leftarrow$ update the imputed matrix using $\textbf{X}^{mis}_{i}$;
        \EndFor
    \EndWhile
\State \Return $\textbf{X}^{new}$
\end{algorithmic}
\end{algorithm}

The stopping criterion is met if the difference between the newly imputed data matrix $\textbf{X}^{new}$ and the imputed data matrix $\textbf{X}^{old}$ from last iteration increases for the first time, which indicates that the imputation process has converged. The difference for the data matrix is defined as:
\begin{equation}
\Delta = \frac{\sum (\textbf{X}^{new} - \textbf{X}^{old})^2}{\sum\textbf{X}^{{new}^2}},
\end{equation}

After imputing the missing values, we assess the performance of imputation using the root mean squared error (RMSE) which is defined by
\begin{equation}
    RMSE = \sqrt{\frac{(\textbf{X}^{com} -  \textbf{X} )^2}{\#NA}}
\end{equation}
where \#NA is the number of missing values in the data matrix.

\section{Experiment}
\label{sec_exp}
In this section, we validate the performance of the proposed algorithm on five public datasets by applying various missingness conditions. We first go through the datasets and baselines, followed by a detailed description of our experiment setup. Then we show imputation and prediction results along with comprehensive analysis.

\subsection{Datasets}
\label{sec_data}
We used the five public datasets for model evaluation:  Messidor \cite{decenciere_feedback_2014}, Spam, COPD \cite{regan2011genetic}, Letter \cite{frey1991letter} and Credit \cite{yeh2009comparisons}, and Table \ref{data} shows the properties of them. Considering that deep learning models are prone to perform well on large and high dimensional datasets, we also include some low dimensional and small size datasets to test the extremes and to prove that our model has real world applications. 

\begin{itemize}
    \item \textbf{Messidor} A dataset of features extracted from medical scans to predict whether a scan contains signs of diabetic retinopathy or not. The image analysis and feature extraction process is described in \cite{antal2014ensemble}.
    \item \textbf{Spam} A dataset of features extracted from e-mail messages to predict whether or not a message is spam. Three features describes the shortest, average, and total length of letters in the message, and the other 54 are values describing the frequency with which certain key words are used.
    \item \textbf{COPD} A dataset of subjects to investigate the genetic susceptibility of chronic obstructive pulmonary disease (COPD). Features include demographic data and medical history, kinds of symptoms questionnaires, medical record review, physical examination, and spirometric measures of lung function.
    \item \textbf{Letter} A dataset of features extracted from distorted images of letters. The objective is to identify each sample as one of the 26 capital letters in the English alphabet. 
    \item \textbf{Credit} A dataset of demographic factors, credit data, history of payment, and bill statements features of credit card clients, to predict whether or not the card holder will default in payment next month.
\end{itemize}

\begin{table*} 
\centering 
\caption{Properties of the datasets. \#Instance: the number of instances in the dataset, \#Continuous: the number of continuous variables, \#Categorical: the number of categorical variables.} \label{data} 
\setlength{\tabcolsep}{5mm}
\begin{tabular}{c|ccc} 
\hline 
\textbf{Dataset} &\textbf{\#Instance} &\textbf{\#Continuous}&\textbf{\#Categorical}\\
\hline 
\hline
Messidor \cite{decenciere_feedback_2014} &1151& 19 & 0\\ \hline
Spam &4601& 57& 0 \\ 
\hline
COPD \cite{regan2011genetic} &4805& 72 & 185\\ 
\hline
Letter \cite{frey1991letter} &20000& 0&16\\ 
\hline
Credit \cite{yeh2009comparisons} &30000& 14&9\\ 
\hline 
\end{tabular} 
\end{table*}  

\subsection{Baselines}
\label{sec_benchmark}
We compare the imputation performance of IFGAN to the following baseline models. 

\begin{itemize}
    \item \textbf{Mean} Mean imputation fills in the missing value $\textbf{X}_i^{mis}$ with the mean of the observed cases $\textbf{X}_i^{obs}$. 
    \item \textbf{KNN}\cite{troyanskaya2001missing} The k-nearest neighbors algorithm firstly finds the $k$ closest neighbors in $\textbf{X}_{\sim i}^{obs}$ to the observation with missing data $\textbf{X}_{\sim i}^{mis}$, and then imputes $\textbf{X}_i^{mis}$ based on the observed cases $\textbf{X}_i^{obs}$ in the neighbors. 
    \item \textbf{SVD}\cite{hastie1999imputing} The SVD method is an iterative method, where in each iteration before convergence, it computes a low rank-$k$ approximation of the data matrix $\textbf{X}$ and replaces the missing values $\textbf{X}^{mis}$ with the corresponding values from the rank-$k$ approximation.
    \item \textbf{MICE}\cite{azur2011multiple} MICE defines a feature-specific imputation process like ours, however, their base models are linear, and are trained in a purely discriminative way. 
    \item \textbf{GAIN}\cite{yoon2018gain} GAIN proposed an adversarial imputation scheme by training on a GAN network on the whole data matrix. The model is adopted to infer $\textbf{X}^{mis}$ by firstly training the generator on the observed cases $\textbf{X}^{obs}$, and then apply the missing mask $\textbf{1} -\textbf{M}$ to obtain corresponding missing entry's value.
\end{itemize}

\begin{table*}
\centering 
\caption{Imputation results comparing our model with five baselines for MCAR setting. Results are displayed using root mean square error(RMSE) with standard deviation. The percentages in the last row indicate the imputation performance improvement of adding the discriminator component. The best result for each dataset is highlighted in boldface.} 
\label{MCAR} 
\begin{tabular}{c|ccccc} 
\hline
\textbf{Method}&\textbf{Messidor}&\textbf{Spam}&\textbf{COPD}&\textbf{Credit}&\textbf{Letter}\\ 
\hline
\hline 
Mean    &0.3317$\pm$ 0.0019 	&0.0573$\pm$0.0011 	&0.3192$\pm$0.0009 	&0.2037$\pm$0.0006 	&0.1905$\pm$0.0005  \\ 
\hline
KNN     &0.1190$\pm$ 0.0057 	&0.0421$\pm$ 0.0012 	&0.2459$\pm$ 0.0006 	&0.1177$\pm$0.0015 	&0.1178$\pm$0.0007 \\ 
\hline
MICE    &0.1042$\pm$ 0.0033 	&0.0473$\pm$ 0.0012 	&0.1752$\pm$ 0.0004 	&0.0986$\pm$ 0.0004 	&0.1107$\pm$0.0003\\ 
\hline
SVD     &0.1472$\pm$0.0044 	&0.0969$\pm$0.0049 	&0.2211$\pm$0.0005 	&0.1487$\pm$0.0011 	&0.1856$\pm$ 0.0013\\ 
\hline
GAIN    &0.3636$\pm$0.0105 	&0.1465$\pm$0.0219 	&0.3616$\pm$0.0005 	&0.3457$\pm$0.0166 	&0.3509$\pm$0.0190\\
\hline
Ours w/o $D$ &0.1025$\pm$0.0031 	&0.0426$\pm$0.0011 	&0.1750$\pm$0.0005 	&0.1080$\pm$0.003 	&0.1092$\pm$0.0008\\
\hline 
\multirow{2}{*}{Ours}  &\textbf{0.0943$\pm$0.0029} &\textbf{0.0412$\pm$0.0013}&\textbf{0.1722$\pm$0.0007}& \textbf{0.0947$\pm$0.0031} &\textbf{0.1071$\pm$0.0006}\\
~& (8\%)& (3\%) & (2\%) & (14\%) & (2\%)\\
\hline 
\end{tabular} 
\end{table*}

\subsection{Experiment Setup}
\label{sec_detail}
To compare the imputation performance under the same metric, one-hot encoding is applied to categorical variables. For continuous variables, we use Min-Max scaler to scale the data to have range from 0 to 1.

We run all the experiments on a server with Intel Xeon E5 CPU and 250G memory. Code is written in Python 3.6. For each feature-specific iteration, the generator is trained for 500 times and the discriminator for 100 times alternately, with learning rate being 0.001. Each experimental run is generated using the stochastic gradient descent optimizer \cite{bottou2010large} with a batch size of 200. The hyper parameters $\alpha$ in the loss function is set to 0.01, which is empirically found to provide good performance in adversarial training. L2 regularization term $\lambda$ is 0.5 in our case.

We conduct each experiment 5 times and report the performance metric along with their standard deviations across the 5 experiments.

\begin{table*}[!h]
\centering 
\caption{Imputation results comparing our model and five baseline models for MAR and MNAR setting. Results are displayed using RMSE with standard deviation. The best result for each dataset is highlighted in boldface.} 
\label{MNAR} 
\begin{tabular}{c|ccccc} 
\hline
\textbf{Method}&\textbf{Messidor}&\textbf{Spam}&\textbf{COPD}&\textbf{Credit}&\textbf{Letter}\\ 
\hline
\hline
\multicolumn{6}{c}{\textbf{Missng At Random (MAR)}}\\
\hline
Mean    &0.3381$\pm$0.0076 	&0.0637$\pm$0.0065 	&0.320$\pm$0.00589 	&0.2068$\pm$0.0046 	&0.1948$\pm$0.0077\\ 
\hline
KNN     &0.1186$\pm$0.0137 	&0.0502$\pm$0.0080 	&0.2515$\pm$0.009 	&0.1220$\pm$0.0097 	&0.1779$\pm$0.0049\\
\hline
MICE    &0.1138$\pm$0.0109 	&0.0479$\pm$0.008 	&0.1855$\pm$0.0013 	&0.1071$\pm$0.0023 	&0.1193$\pm$0.0012\\
\hline
SVD     &0.1515$\pm$0.0135 	&0.0914$\pm$0.0020 	&0.2228$\pm$0.0084 	&0.1541$\pm$0.0040 	&0.1937$\pm$0.0032\\
\hline
GAIN    &0.4337$\pm$0.0228 	&0.1207$\pm$0.0282 	&0.4239$\pm$0.0067 	&0.3563$\pm$0.0226 	&0.4202$\pm$0.0121\\
\hline
Ours    &\textbf{0.0886$\pm$0.0105} &\textbf{0.0457$\pm$0.0032}&\textbf{0.1815$\pm$0.0044} & \textbf{0.0938$\pm$0.0080} &\textbf{0.1147$\pm$0.0067}\\
\hline 
\multicolumn{6}{c}{\textbf{Missng Not At Random (MNAR)}}\\
\hline
Mean    & 0.3314$\pm$ 0.0112 	&0.0608$\pm$0.0124 	&0.3261$\pm$0.0065 	&0.2135$\pm$0.0126 	&0.1976$\pm$0.0159 \\
\hline
KNN     & 0.1197$\pm$0.0149 	&0.0496$\pm$0.0144  	&0.2519$\pm$0.0112 	&0.1215$\pm$0.0138 	&0.1801$\pm$0.0131\\ 
\hline
MICE    & 0.1107$\pm$0.0168 	&0.0485$\pm$0.0183 	&0.1832$\pm$0.0094 	&0.1056$\pm$0.0032 	&0.1143$\pm$0.0073\\ 
\hline
SVD     & 0.1498$\pm$0.0173 	&0.0967$\pm$0.0116 	&0.2304$\pm$0.0130 	&0.1545$\pm$0.0121 	&0.1892$\pm$0.0072\\
\hline
GAIN    & 0.4415$\pm$0.0156 	&0.1280$\pm$0.0258 	&0.4230$\pm$0.0067 	&0.3463$\pm$0.0128 	&0.4207$\pm$0.0074\\ 
\hline
Ours    &\textbf{0.0879$\pm$0.0180} &\textbf{0.0476$\pm$0.0093} &\textbf{0.1818$\pm$0.0065}& \textbf{0.0892$\pm$0.0132} &\textbf{0.1122$\pm$0.0118}\\
\hline 
\end{tabular} 
\end{table*}

\subsection{Imputation Performance}
\label{sec_imp}
In order to test IFGAN's robustness in different missing settings, we vary the missing mechanism, missing type, missing rate, feature size and sample size respectively and report the results. Due to the limitation of space, we show the missing type variations experiments on all five datasets, and the missing rate, feature size and sample size variations experiments on the Credit dataset.

\textbf{Missing Mechanisms.} We firstly apply different types of missingness to the datasets by taking the following steps: 
\begin{enumerate}
    \item Generate a uniform random matrix $\textbf{Z}$ with values between 0 and 1 of $N$ observations and $d$ dimension.
    \item Generate (the data matrix with missing values $\hat{\textbf{X}}$ and the mask matrix $\textbf{M}$) using (\textbf{Z}, the original data matrix \textbf{X} and the missing rate $t$) according to the following three missing data mechanisms:
    \begin{enumerate}
    \item[a)] \textit{MCAR}: Set all entries in $\textbf{X}^{com}$ to have missing values if corresponding  $\textbf{Z}_{ij} \leq t$, where $i \in 1:N, j \in 1:d$.
    \item[b)] \textit{MAR}: Randomly sample one feature $\textbf{X}^{com}_{r}$ from the data matrix and calculate its median $m_{r}$. Set ten randomly sampled features $\textbf{X}^{com}_{a1}, \cdots, \textbf{X}^{com}_{a10}$ from all features except $\textbf{X}_{r}$ to have missing values where $\textbf{Z}_{ij} \leq t$ and $\textbf{X}^{com}_{ir} \leq m_r$, for $i \in 1:N, j \in \{a_1,...,a_{10}\}$. Set all other features except $\textbf{X}^{com}_{a_1}, \cdots, \textbf{X}^{com}_{a_{10}}$ to have missing values where $\textbf{Z}_{ij} \leq t$, for $i \in 1:N, j \in \{1:d\} \backslash \{a_1, \cdots,a_{10}\}$. For the experiments on this missing mechanism, we set $t$ equal to 0.5.
    \item[c)] \textit{MNAR}: Since MNAR has a greater effect on the data distribution, we randomly sample five features $\textbf{X}^{com}_{a_1}, \cdots, \textbf{X}^{com}_{a5}$ from the data matrix and calculate their median $m_{a_1}, \cdots, m_{a_5}$ separately. Set these features to having missing values where $\textbf{Z}_{ij} \leq t$ and $\textbf{X}^{com}_{ij} \leq m_j$, where $i \in 1:N, j \in \{a_1, \cdots, a_5\}$. Set all other features except $\textbf{X}^{com}_{a_1}, \cdots, \textbf{X}^{com}_{a_5}$ to have missing values where $\textbf{Z}_{ij} \leq t$, for $i \in 1:N, j \in \{1:d\} \backslash \{a_1, \cdots, a_5\}$. For experiments on this missing mechanism, we set $t$ equal to 0.5.
    \end{enumerate}
\end{enumerate}

We firstly apply different types of missingness to the datasets by taking generating missing entries according to Missing Completely At Random (MCAR), Missing At Random (MAR) and Missing Not At Random (MNAR) settings.

We report the imputation performance on real life datasets with MCAR missing in Table \ref{MCAR}, by comparing the imputation results (in RMSE) obtained by our model with five baseline models. As can be seen from the table, IFGAN significantly outperforms other benchmarks in all cases, especially when the datasets' sample size is large(e.g. Credit and Letter) or feature dimension is high(e.g. COPD).
As shown in Table \ref{MCAR}, we also exclude the discriminator and compare the performance of the resulting architecture against our original setting. The performance dropped by 2\% $\sim$ 17\% when the discriminator is removed, indicating that our framework benefits from the adversarial training process.

scFor the MAR and MNAR (in Table \ref{MNAR}) missing settings, which are bottlenecks for most imputation algorithms, IFGAN also yields promising results. Our model is capable of dealing with non-random missing data generation processes because we don't place strict assumptions about the distribution of the data or subsets of the variables. The only exception is the Spam dataset with MNAR missing mechanism, and in this case, KNN has a better performance. This is because KNN imputer is designed for continuous data and all features in Spam are continuous. However, KNN cannot handle categorical and mixed type data, and fails to compete with our model under these conditions.


\begin{table*}[!h]
\centering 
\caption{Prediction performance in terms of Area Under the Receiver Operating Curve (AUROC) on Credit dataset. We run the experiment with 5-fold cross validation. The best result for each missing rate is highlighted in boldface.} 
\label{tab_class} 
\begin{tabular}{c|ccccccc} 
\hline
\textbf{Method}&\textbf{0.1}&\textbf{0.2}&\textbf{0.3}&\textbf{0.4}&\textbf{0.5}&\textbf{0.6}&\textbf{0.7}\\ 
\hline 
Mean	&0.7637 &0.7381 &0.7282 &0.7251 &0.7189 &0.6990 &0.7041\\
KNN	  &0.7634 &0.7486 &0.7393 &0.7175 &0.7104 &0.6911 &0.6859  \\ 
MICE    &0.7683 &0.7566 &0.7497 &0.7244 &0.7145 &0.7152 &\textbf{0.7150}\\ 
SVD	 &0.7517 &0.7433 &0.7386 &0.7337 &0.7111 &0.6917 &0.6923 \\ 
GAIN	&0.5052 &0.4770 &0.5146 &0.5060 &0.4948 &0.4952 &0.4981  \\ 
\hline
Ours	 &\textbf{0.7766} 	&\textbf{0.7786} 	&\textbf{0.7598} 	&\textbf{0.7493} 	&\textbf{0.7371} 	&\textbf{0.7266} 	&0.7107 \\
\hline 
\end{tabular} 
\end{table*}

\begin{figure}[!h]
\centering
	\centering 
	\includegraphics[width=0.8\linewidth]{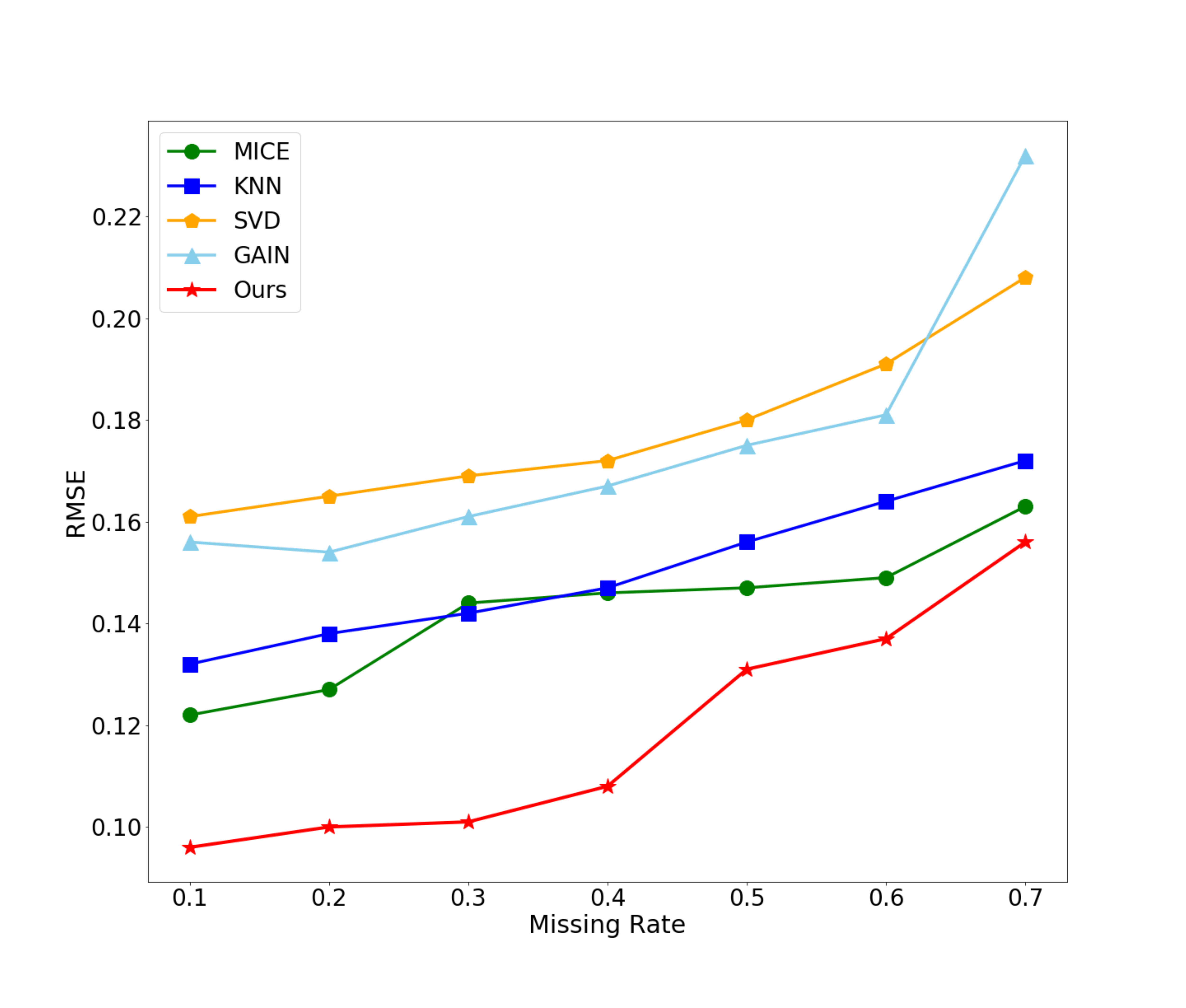}
\caption{The imputation performance on Credit with varying missing rate from 10\% to 70\%.}
\label{vary_missing_rate} 
\end{figure}

\begin{figure}[!h]
\centering
	\centering 
	\includegraphics[width=0.8\linewidth]{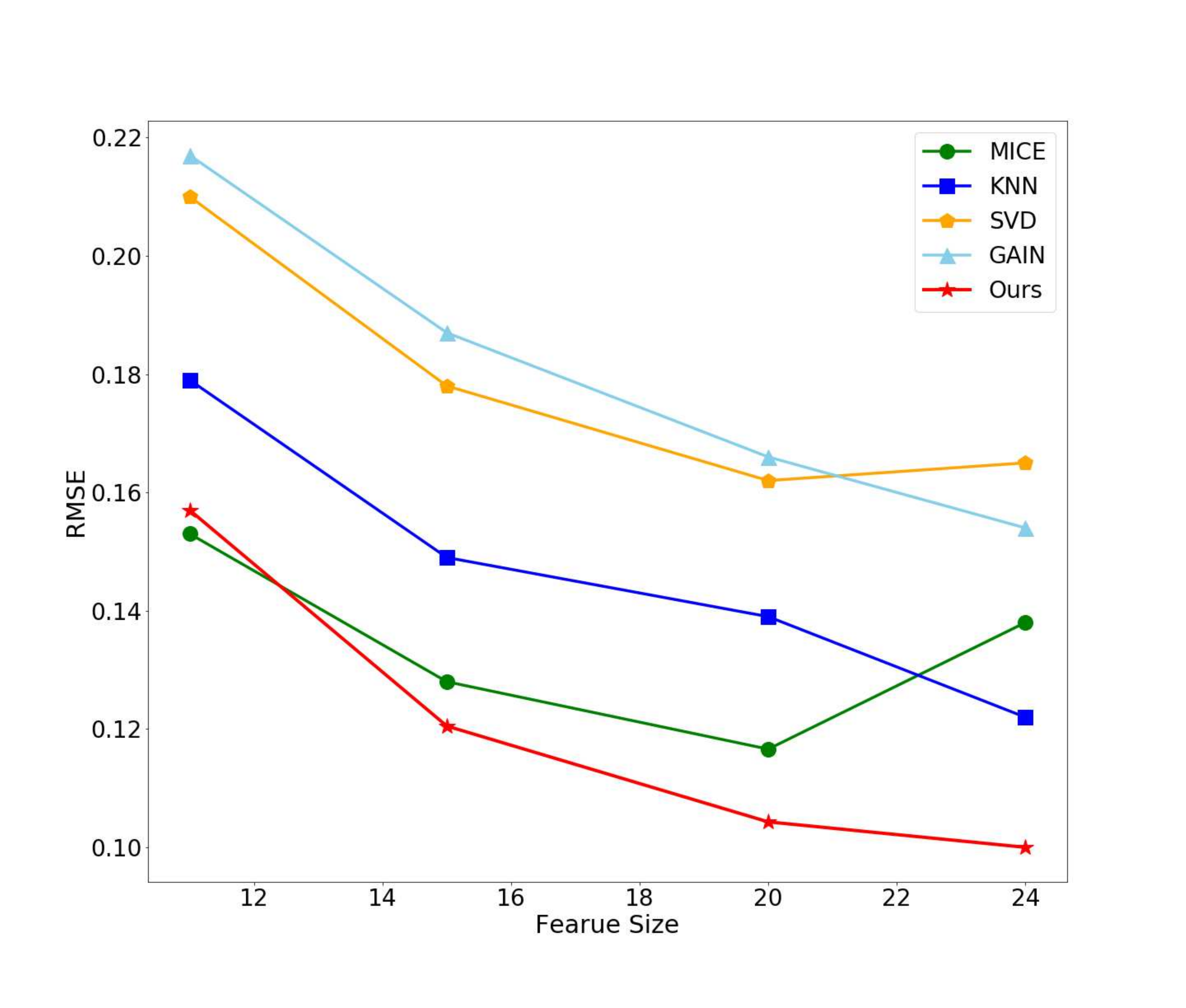}
\caption{The imputation performance on Credit with varying feature size. Missing rate is set to 0.2.} 
\label{vary_feature_size} 
\end{figure}

\begin{figure}[!h]
\centering
	\centering 
	\includegraphics[width=0.8\linewidth]{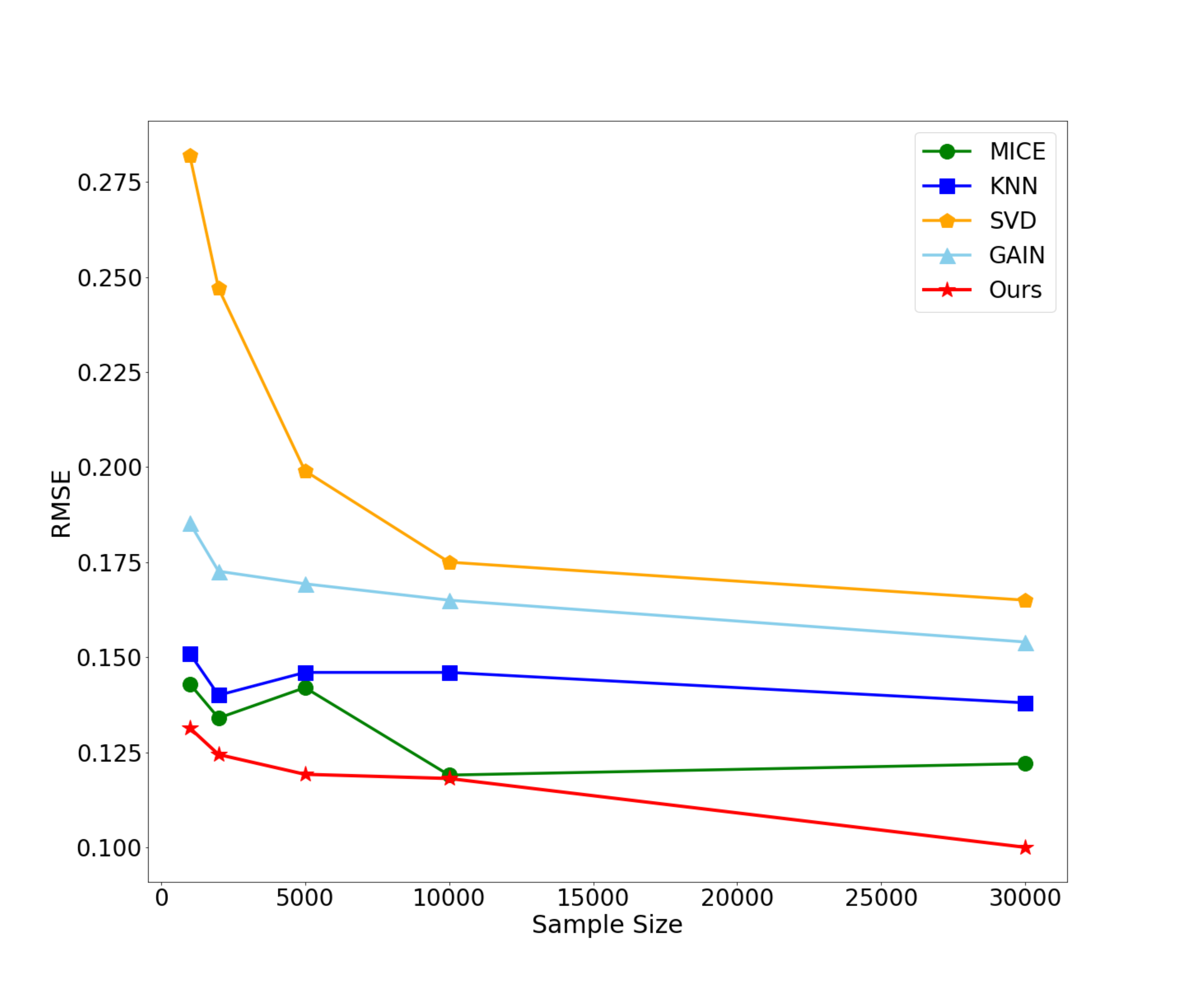}
\caption{The imputation performance on Credit with varying sample size. Missing rate is set to 0.2.}
\label{vary_sample_size} 
\end{figure}

\textbf{Missing Rate.} 
Missing data proportion is known to affect imputation performance, which deteriorates with increasing missing data proportion. In order to test the impact of different missing data proportion on our model, we introduce missingness to the Credit dataset by varing the missing rate from 10\% to 70\% with the stride equal to 10\%. The result in Figure \ref{vary_missing_rate} indicates that although the performance drops slightly when the missing proportion rises, IFGAN still outperforms other baselines, showing its robustness across the entire range of missing rates.

\textbf{Feature size.} We also test our model against different feature sizes by sampling a subset of features from the Credit dataset. Figure \ref{vary_feature_size} shows that although IFGAN use the deep learning architecture as its backbone, it can handle low dimensional data input as well. It is worth noting that, as the feature size increases, the performance improvements of IFGAN over the linear model MICE also increases.

\textbf{Sample size.}
Imputation models are also sensitive to sample size. As illustrated in Figure \ref{vary_sample_size}, IFGAN is still able to outperform other methods when the number of samples is relatively small, while matrix completion method like SVD cannot handle this situation. Even though the performance of each algorithm increases as sample size rises, our model consistently has the best imputation performance compared with baselines.

\begin{figure}[h]
    \centering
    \includegraphics[width=0.8\linewidth]{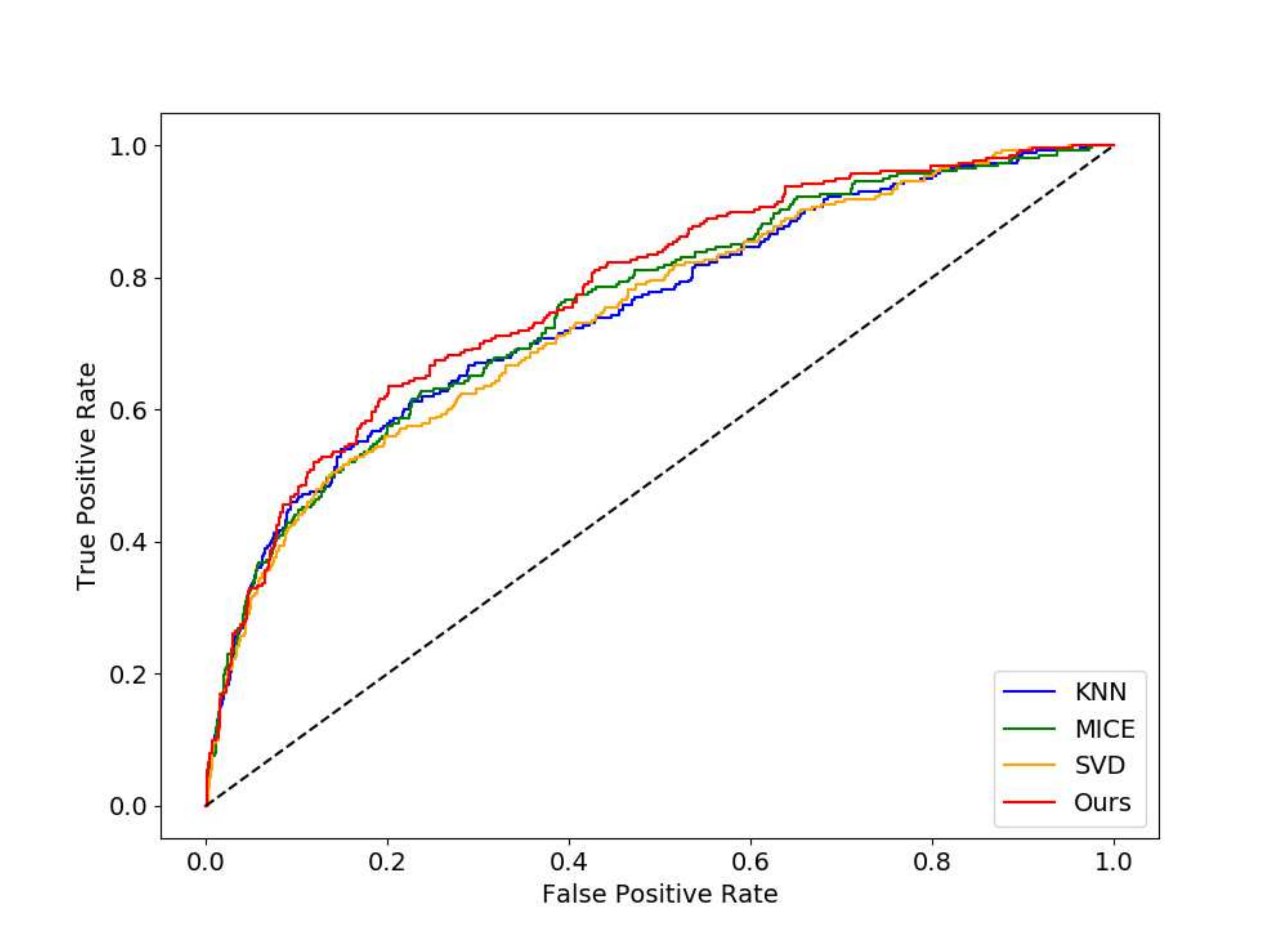}
    \caption{The Receiver Operating Curve of the classification result using the dataset imputed by different methods with missing rate set to 0.2. }
    \label{fig_roc}
\end{figure}

\subsection{Post-imputation performance}
\label{sec_post}
The goal of imputation is to generate complete datasets which can be used for knowledge discovery and analysis. While imputation accuracy provides us with a measure of how close the imputed dataset $\textbf{X}$ is to the original complete dataset $\textbf{X}^{com}$, we still do not know how well inter-variable correlations are preserved, which will greatly affects the quality of data analysis. Taking this into consideration, we used the imputed datasets obtained by each benchmark to test the the imputation quality with regards to the preservation of a dataset’s overall structure, and to quantify the impact of imputation on post-imputation analysis.

Since the test dataset (Credit) is used for the classification task, we adopt Area Under the Receiver Operating Curve (AUROC) as the measure of performance. The classifier we use is XGBoost \cite{chen2016xgboost} as the classifier, and the parameter tuning technique is random search\cite{bergstra2012random} implemented by scikit-learn machine learning package\cite{scikit-learn}. In the experiment, we used various missing rates to test the consistency of results.

As Table \ref{tab_class} and Figure \ref{fig_roc} show, our method, which achieved the best imputation accuracy, yields the best post-imputation prediction accuracy in most missing rate cases. This indicates that using our model for imputation provides higher predictive power for post-imputation analysis compared to other benchmarks.

\section{Conclusion and Future Work}
\label{sec_con}

We have presented IFGAN, a novel method for missing data imputation in this paper. Our model is based on a feature-specific generative adversarial network and is capable of handling various conditions of missingness. The experiments on several real-life datasets demonstrate that IFGAN outperforms current state-of-the-art algorithm against different missing mechanisms, missing rates, data types, feature sizes, and sample sizes. Besides, the proposed model neither requires much effort for parameter tuning nor needs a complete dataset to train on. It also improves post-imputation analysis by preserving inter-variable correlations during learning.

Future work will investigate clinical health records \cite{wells2013strategies} where MAR and MNAR missingness often take place. The performance of IFGAN on datasets of a certain domain is also worth studying. For instance, how to incorporate the domain knowledge and priors into the imputation pipeline may be crucial to the quality of synthetic data.

\bibliography{refer.bib}
\bibliographystyle{IEEEtran}

\end{document}